\documentclass{preprint}

\usepackage{graphicx}
\usepackage{booktabs}
\usepackage{amsmath}
\usepackage{multirow}
\usepackage{subcaption}

\begin{document}

\articletype{Paper}

\title{Learning Control as Enabling Layer for Embodied Intelligence Research explored with Soft Robotic Swimming in diverse Flow Speeds}

\author{Fabian Schwab$^{1,2}$, Federico Allione$^1$, Bingcheng Wang$^{1,2}$, Mohamed El Arayshi$^3$, Claudio Mucignat$^1$, Ivan Lunati$^1$, Cristiano M. Verrelli$^3$ and Ardian Jusufi$^{1,2,*}$}

\affil{$^1$Engineering Sciences Department, Swiss Federal Laboratories for Materials Science and Technology, Switzerland}
\affil{$^2$Institute for NeuroInformatics, University of Zurich, Switzerland}
\affil{$^3$Department of Electronic Engineering, University of Rome Tor Vergata, Italy}

\email{ardian@ini.uzh.ch}


\begin{abstract}

Soft robots are valuable robophysical platforms for studying body-caudal undulatory locomotion, but their compliant bodies are difficult to control precisely under changing hydrodynamic loading. Conventional proportional--integral--derivative (PID) feedback stabilises periodic undulation in static water but accumulates a flow-dependent tracking delay and increasing inter-trial variability when the environmental flow is non-trivial, due to the nonlinearities of the system. Here, we evaluate whether augmenting PID with a Linear Repetitive Learning Estimation Scheme (PID-LRLES) recovers tracking accuracy and repeatability under dynamic flow. Such LRLES generalizes the classical integral action to the case of periodic (non-constant) references, with the resulting learning scheme being constituted by a transfer function with all its poles with
negative real part, to avoid the typical long-term instability issues of classical repetitive controls. Closed-loop experiments were carried out in a recirculating flow tank at five bulk flow speeds spanning $0$ to $32.6~\mathrm{cm\,s^{-1}}$, using an embedded soft capacitive bending sensor at the $1~\mathrm{kHz}$ control-loop rate. With controller gains tuned once in static water and held fixed across all conditions, PID-LRLES tracked the periodic bending-envelope reference more closely than the PID baseline and significantly reduced the inter-trial spread of the per-trial RMSE (paired Wilcoxon signed-rank test, $p = 1.8 \times 10^{-4}$, $n_{\mathrm{pairs}} = 25$). Embedded soft proprioception and cycle-to-cycle learning act as complementary contributions to robustness: the sensor exposes the periodic hydrodynamic bias in the body deformation, while the learning term absorbs it over the past few oscillation cycles. By reducing flow-dependent control-induced variability, the approach provides an enabling layer for future robophysical studies seeking to isolate the effects of morphology, sensing and environmental flow on aquatic locomotion.

\end{abstract}

\section{Introduction}

Aquatic animals locomote through complex and dynamically changing fluid environments with a robustness and efficiency that engineered systems still struggle to match~\cite{dickinson2000animals,Ijspeert2020}. Their swimming performance emerges from the interplay of body morphology, distributed sensing, neural control and surrounding hydrodynamics: flexible anatomies combined with integrated proprioception endow animals with a form of `morphological intelligence' that supports stable, adaptive movement under perturbation~\cite{Woodward2018,Siddall2019}. Within this framework, the biological motor system separates motor commands that tune the body --- setting limb stiffness and configuration to suit the task (motor commands for tuning, MCT) --- from motor commands that directly drive movement and interaction forces (MCM); the effectiveness of the latter depends on how well the former is regulated, and soft-bodied robots are particularly natural platforms on which to study this coupling~\cite{Nanayakkara2024}. Replicating such behaviour in robots remains challenging, however, because compliant bodies introduce nonlinear deformation, distributed dynamics and strong fluid--structure coupling that are difficult to model precisely.

Soft robotics offers a route to emulate these properties using compliant, bio-mimetic materials~\cite{Ijspeert2020}, and aquatic locomotion has become a particular focus because fish combine high efficiency with agility in unsteady flow~\cite{Lauder2007,Smits2019}. Fish modulate the amplitude and frequency of their body--caudal undulation and exploit the passive stiffness of their bodies to extract energy from the surrounding fluid~\cite{McHenry1995,Lauder2011}, making caudal-fin oscillation one of the most cost-effective propulsion modes~\cite{Block1992,Liao2003}. These principles have inspired a range of soft robotic swimmers spanning body--caudal undulators~\cite{Barrett1996,Marchese2014,Katzschmann2018,Aubin2019,Struebig2020} and pectoral-fin oscillators that exploit snap-through actuation to reach high speed and manoeuvrability with simple pneumatic inputs~\cite{Qing2024}; however, passive or open-loop designs struggle to maintain consistent kinematics when flow velocity or actuation frequency change.

\begin{figure}[!ht]
    \centering
    \begin{subfigure}[t]{0.48\linewidth}
        \centering
        \includegraphics[width=\linewidth]{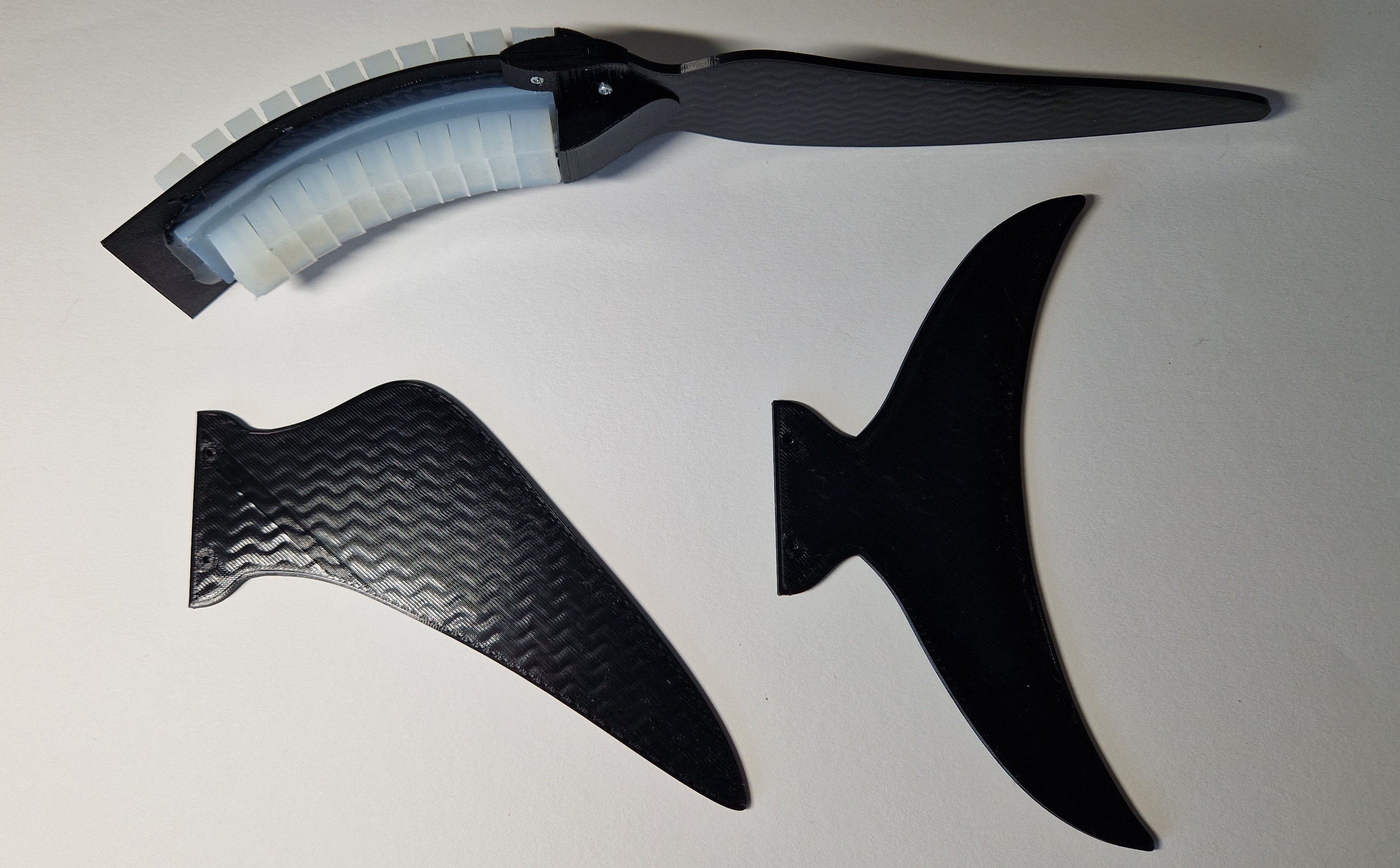}
        \caption{}
        \label{fig:fins_torque_a}
    \end{subfigure}
    \hfill
    \begin{subfigure}[t]{0.48\linewidth}
        \centering
        \includegraphics[width=\linewidth]{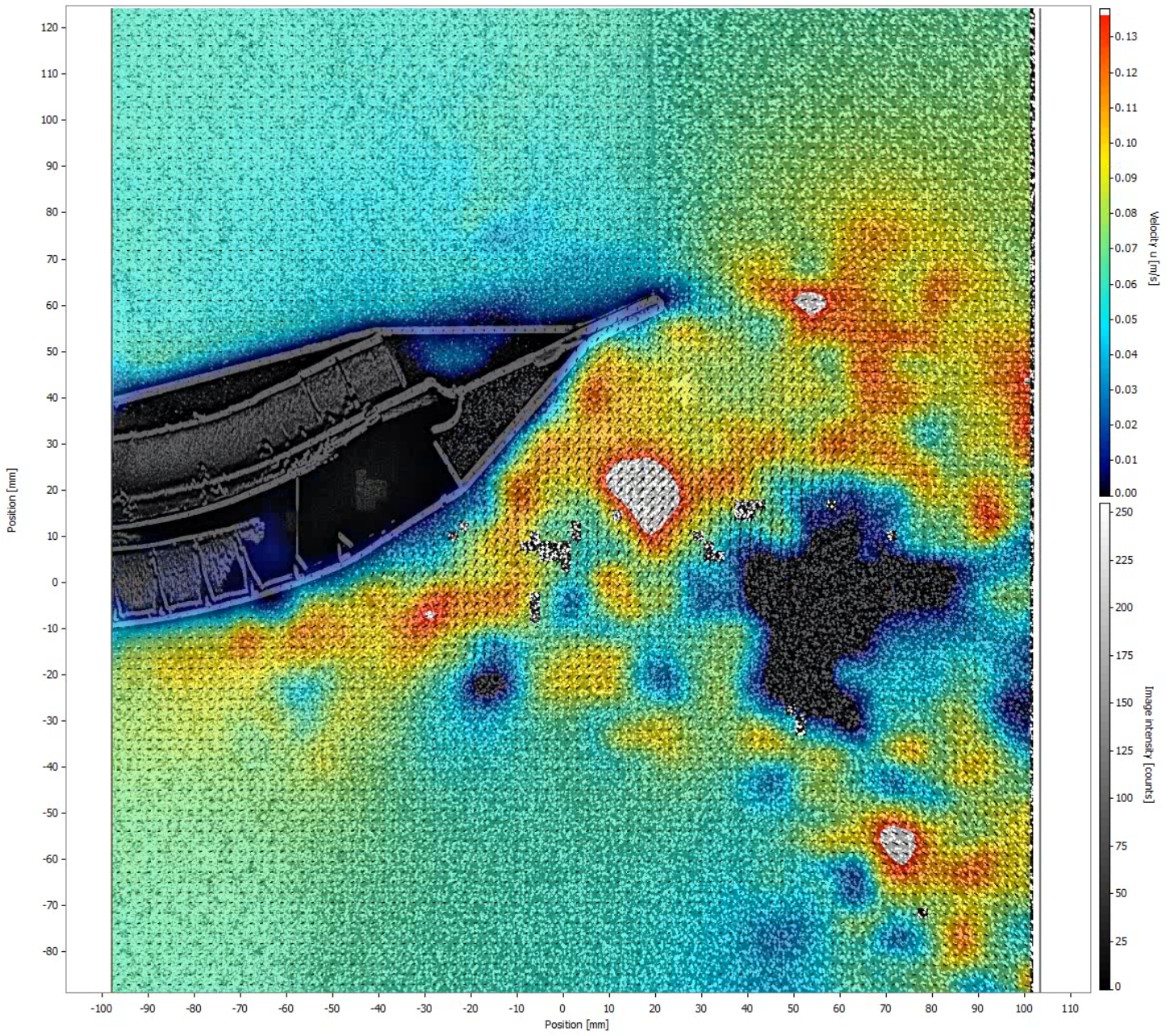}
        \caption{}
        \label{fig:fins_torque_b}
    \end{subfigure}

    \vspace{0.5em}

    \begin{subfigure}[t]{0.7\linewidth}
        \centering
        \includegraphics[width=\linewidth]{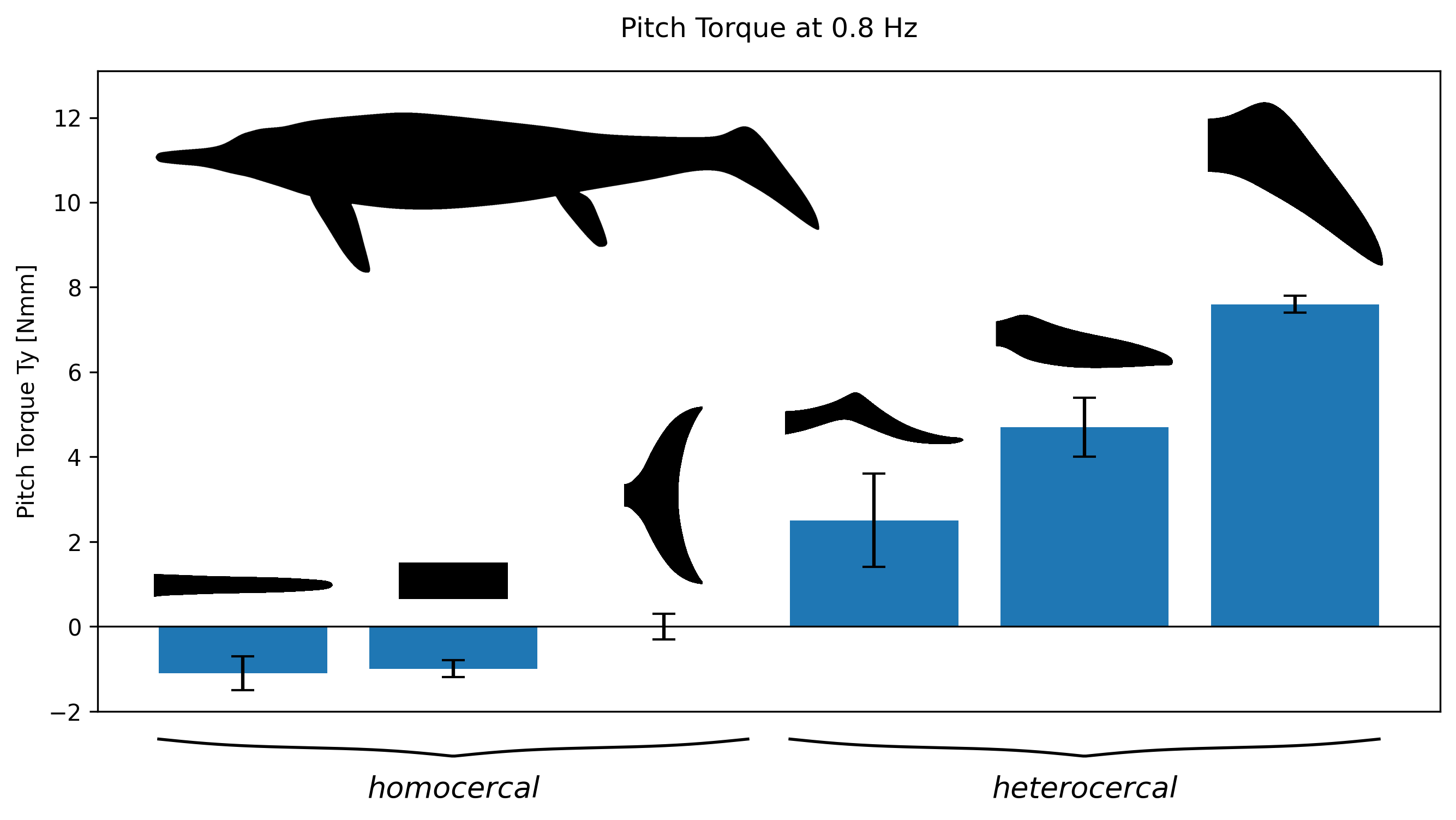}
        \caption{}
        \label{fig:fins_torque_c}
    \end{subfigure}
    \caption{
    Form and function relationships arising from shape and control.
    (a) Family of extinct ichthyosauriform caudal-fin morphologies that can be mounted on the platform's detachable tail cuff, photographed against the soft pneumatic body--caudal swimmer; the inset schematic indicates the thrust $F$, vertical thrust and pitch torque conventions used in the rest of the paper.
    (b) Particle image velocimetry (PIV) snapshot of the velocity-magnitude field around the soft robotic fish under flow in the Empa recirculating flow tank; the colour scale denotes velocity magnitude in $\mathrm{m\,s^{-1}}$. Reproduced from the PIV characterisation of the platform reported in~\cite{Schwab2022}, included here to illustrate the hydrodynamic wake structure that mediates between caudal-fin shape and the measured pitch torque shown in panel~(c).
    (c) Pitch torque generated by these caudal fins at $f_{\mathrm{flap}} = 0.8~\mathrm{Hz}$, ordered from homocercal (left) to heterocercal (right) tail-fin shapes. Data reproduced and re-rendered from~\cite{Sprumont2024} with permission of IOP Publishing, to illustrate the kind of morphology-dependent robophysical inference that the present work is designed to support.
    }
    \label{fig:fins_torque}
\end{figure}

Beyond their engineering appeal, soft robotic fish serve as robophysical models: controllable physical platforms for testing biomechanical hypotheses that are impractical to probe in live animals~\cite{Nyakatura2019,siddall2021b}. We have previously characterised such a platform hydrodynamically using particle image velocimetry~\cite{Schwab2022}, instrumented it with soft strain sensors to compare body--caudal undulation in robots and live fish~\cite{Schwab2021}, and used it to show how caudal-fin morphology governs thrust and pitch torque across extinct ichthyosauriform tail shapes~\cite{Sprumont2024} (Figure~\ref{fig:fins_torque}); fin properties such as stiffness can likewise be tuned to improve thrust and efficiency~\cite{Obayashi2025}. The scientific value of these experiments hinges on the repeatability of the generated body--caudal kinematics: if control-induced variability is high, observed differences in thrust, pitch torque or swimming stability cannot be confidently attributed to morphology or environmental flow.

Conventional PID feedback can stabilise the periodic body--caudal undulation needed for repeatable experiments, but it does not exploit the periodic structure of the reference (passing through the system nonlinearities) and its performance is sensitive to changes in hydrodynamic loading. Learning-based control has emerged as a particularly suitable complement to PID for soft robots, where uncertain parameters and unmodelled dynamics make analytical models hard to obtain~\cite{Laschi2025}. Within this broader class, repetitive learning control leverages the periodicity of the reference directly, using error information from previous cycles to refine future actuation commands and thereby compensate for recurring disturbances~\cite{xu2006repetitive,verrelli2022pi}. We previously realised this for a soft robotic fish by augmenting a PID controller with a Linear Repetitive Learning Estimation Scheme (PID-LRLES)~\cite{tomei2015linear,Schwab2024}, which improved body--caudal trajectory tracking but was validated only under static-water conditions. Such LRLES generalized the classical integral action to the case of periodic (non-constant) references, with the resulting learning scheme being constituted by a transfer function with all its poles with negative real part: the typical long-term instability issues of classical repetitive controls were avoided. Indeed, in a preliminary conference study, we extended the evaluation to a recirculating flow tank and found that PID-LRLES reduced tracking-error variability across flow speeds~\cite{SchwabAMAM2025}. It remained open, however, whether this advantage holds as a thorough, statistically supported result across hydrodynamic conditions, and what role embedded proprioceptive feedback plays in adaptation under unsteady flow.

Here, we present a full evaluation of PID-LRLES under controlled dynamic flow, closing the loop on body--caudal deformation with embedded soft proprioceptive feedback. Building on our preliminary flow-tank experiments~\cite{SchwabAMAM2025}, we show that the learning control improves not only tracking accuracy but also the robustness and repeatability of robophysical swimming experiments under varying hydrodynamic conditions, providing an enabling layer for morphology-dependent studies of aquatic locomotion.

The main contributions of this work thus are:

\begin{description}
    \item{i)} validation of PID-LRLES learning control under dynamic flow conditions;
    \item{ii)} integration of soft proprioceptive sensory feedback for closed-loop body--caudal undulation control;
    \item{iii)} demonstration of reduced tracking variability compared with conventional PID control;
    \item{iv)} positioning of learning-enhanced closed-loop control as an enabling layer for repeatable robophysical swimming experiments.
\end{description}

Table~\ref{tab:benefits} summarises the engineering benefits of the proposed approach (soft proprioception + PID-LRLES) relative to a conventional fixed-gain PID baseline across five performance-relevant dimensions, framed against the corresponding biological benchmark in each case.

\begin{table}[!ht]
    \centering
    \footnotesize
    \caption{Engineering benefits of the proposed approach --- embedded soft proprioception combined with PID-LRLES learning control --- relative to a conventional fixed-gain PID baseline, organised by dimension. Numerical entries are taken from sections~\ref{sec:variability} and~\ref{sec:robustness}.}
    \label{tab:benefits}
    \begin{tabular}{p{0.11\linewidth} p{0.18\linewidth} p{0.21\linewidth} p{0.21\linewidth} p{0.18\linewidth}}
        \toprule
        Dimension & Biological benchmark & Conventional baseline (fixed-gain PID) & Proposed approach (PID-LRLES + embedded proprioception) & Engineering benefit \\
        \midrule
        Tracking accuracy
        & Continuous in-cycle adaptation of body--caudal kinematics
        & Persistent envelope delay and amplitude deficit; deviation grows from $\sim 0.02$ to $\sim 0.04~\mathrm{rad}$ across $U \in [0, 32.6]~\mathrm{cm\,s^{-1}}$
        & Cycle-to-cycle learning absorbs the periodic hydrodynamic bias; envelope tracks reference within $\pm 0.015~\mathrm{rad}$
        & $\sim\!40\%$ lower mean RMSE pooled across flow speeds \\
        \addlinespace
        Flow robustness
        & Distributed sensing supports in-situ adjustment to ambient flow
        & Mean RMSE rises monotonically from $0.025$ to $0.034~\mathrm{rad}$ over $U \in [0, 32.6]~\mathrm{cm\,s^{-1}}$
        & Mean RMSE rises only from $0.015$ to $0.018~\mathrm{rad}$ over the same range
        & Markedly flatter flow-speed sensitivity \\
        \addlinespace
        Inter-trial repeatability
        & Low trial-to-trial drift under repeated identical reflex commands
        & Tracking-error distribution widens substantially with flow speed
        & Learning term removes the recurring drift component of the disturbance
        & Significantly reduced inter-trial spread ($p = 1.8 \times 10^{-4}$, paired Wilcoxon, $n_{\mathrm{pairs}} = 25$) \\
        \addlinespace
        Tuning effort
        & Continuous adaptation in living tissue
        & Gains must be retuned per operating point to preserve performance
        & Gains tuned once in static water and held fixed across all flow conditions
        & Removes per-flow recalibration burden \\
        \addlinespace
        Sensing modality
        & Embedded proprioception (e.g.\ muscle spindles, lateral-line organs)
        & Open-loop actuation or external-camera command; flow-induced bias not observable to controller
        & Embedded capacitive bending sensor exposes true body deformation at $1~\mathrm{kHz}$
        & Flow-induced periodic bias appears in $\tilde q(t)$ and is therefore correctable \\
        \bottomrule
    \end{tabular}
\end{table}

\begin{figure}[!t]
    \centering
    \begin{subfigure}[t]{0.48\linewidth}
        \centering
        \includegraphics[width=\linewidth]{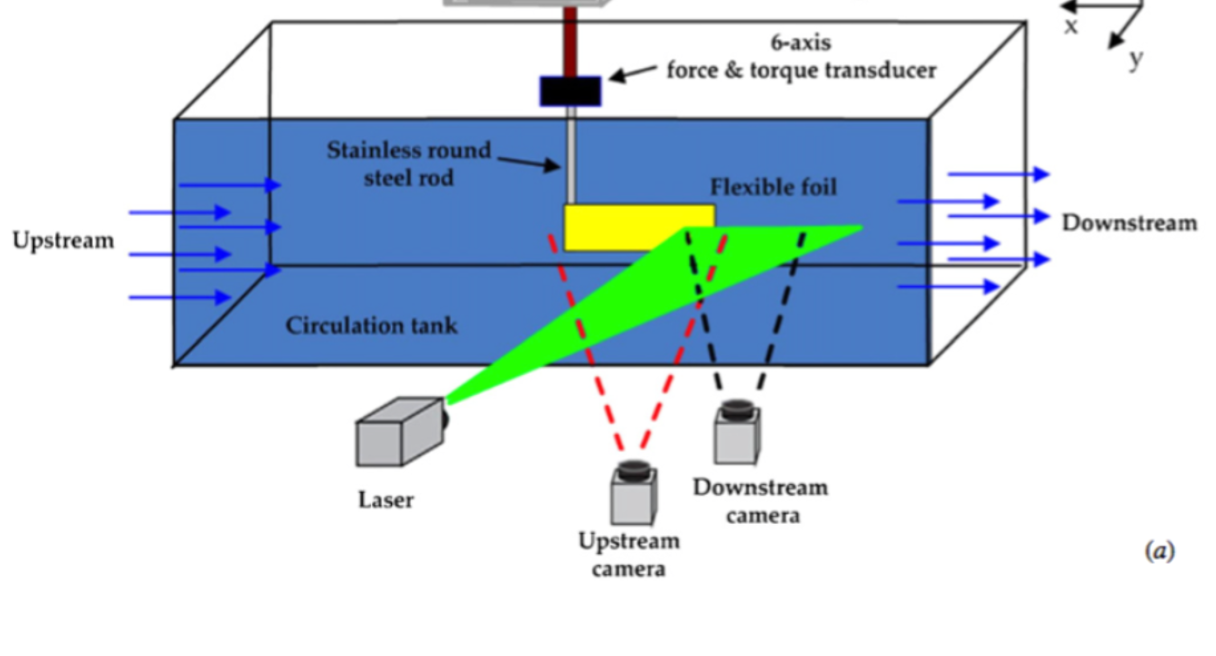}
        \caption{}
        \label{fig:setup_a}
    \end{subfigure}
    \hfill
    \begin{subfigure}[t]{0.48\linewidth}
        \centering
        \includegraphics[width=\linewidth]{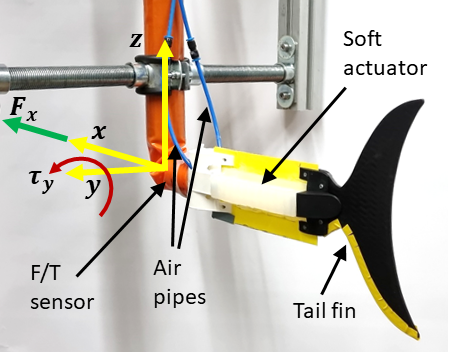}
        \caption{}
        \label{fig:setup_b}
    \end{subfigure}

    \vspace{0.5em}

    \begin{subfigure}[t]{\linewidth}
        \centering
        \includegraphics[width=0.9\linewidth]{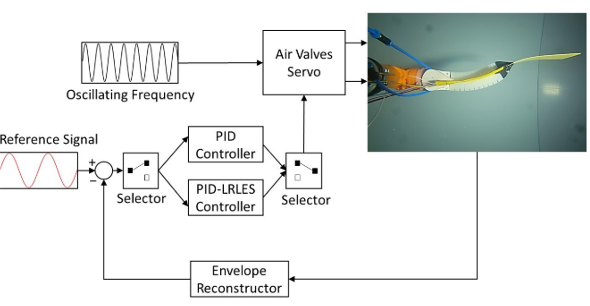}
        \caption{}
        \label{fig:setup_c}
    \end{subfigure}
    \caption{
    Experimental setup and closed-loop control architecture.
    (a) Side view of the recirculating flow tank with the six-axis force/torque transducer above the test section and the below-tank laser and upstream/downstream cameras used for kinematic tracking.
    (b) Close-up of the soft robotic fish with the body-frame axes ($x$, $y$, $z$; forces $F_x$ and pitch torque $\tau_y$) and the embedded soft pneumatic actuator, air pipes, F/T sensor and detachable tail fin labelled. Reproduced from~\cite{Sprumont2024} with permission of IOP Publishing.
    (c) Block diagram of the closed-loop PID and PID-LRLES controllers: the reference envelope and oscillating-frequency commands feed into the selector that switches between the PID and PID-LRLES blocks; the resulting pressure command drives the air-valves servo and hence the soft body--caudal undulation, while the embedded soft sensor signal feeds the envelope reconstructor back into the error path. Reproduced from~\cite{Schwab2024} under the CC~BY~4.0 licence.
    }
    \label{fig:setup}
\end{figure}

The robotic fish platform used in this study (Figure~\ref{fig:setup}) is based on the prototype demonstrated in static-water conditions~\cite{Schwab2024} and characterized hydrodynamically with particle image velocimetry~\cite{Schwab2022}.

\section{Robophysical Fish Platform}
\label{sec:platform}

 It consists of a compliant body--caudal structure actuated by paired soft pneumatic muscles and instrumented with an embedded soft bending sensor that closes the proprioceptive control loop. The fabrication approach follows~\cite{Jusufi2017a} and is based on the soft-actuator design of~\cite{Mosadegh2014}.

\subsection{Compliant body and soft pneumatic actuation}
\label{sec:body}

The robotic body is built around a flexible plastic foil that plays the role of the backbone, to which two soft pneumatic actuators are bonded on opposite sides to provide lateral bending. A frontal cuff (3D-printed, ABS) attaches the body to a fixed mast in the flow tank, while a flexible tail cuff (3D-printed, TPU~A95) holds a detachable passive caudal fin; both cuffs also streamline the body profile. The backbone foil is a $0.52~\mathrm{mm}$-thick shim stock (Artus, Inc.) with Young's modulus $1240~\mathrm{MPa}$ and flexural stiffness $9.9 \times 10^{-4}~\mathrm{N\,m^2}$, comparable to that reported for similarly sized live fish~\cite{Jusufi2017a}. The two actuators, each $10~\mathrm{cm}$ long and $2~\mathrm{cm}$ in height, are cast from a silicone elastomer (Dragon Skin\textsuperscript{TM} 20, Smooth-On Inc.) and bonded to the backbone with a silicone adhesive (Dowsil~734 Flowable Sealant, DOW). The truncated TPU~A95 caudal fin used in the present experiments extends the body by a further $13~\mathrm{cm}$~\cite{Sprumont2024}, giving an overall body length $L_b \approx 0.23~\mathrm{m}$. Alternating pressurisation of the two actuators generates the lateral body--caudal undulation.

Actuation pressure is delivered by a digital pressure regulator (ITV0050-3BS, SMC) operating over the actuator's $0.7$--$2.5~\mathrm{bar}$ working range. Each actuator is connected via two solenoid valves (SYJ7320-5LOU-01F-Q, SMC) that switch its volume between the regulated supply line and atmosphere; an intermediate air tank buffers the supply against transient pressure drops on actuator venting.

\subsection{Embedded soft proprioceptive feedback}
\label{sec:sensor}

A soft capacitive bending sensor (1-Axis Soft Flex Sensor, Nitto Bend Technologies) is affixed to a flexible foil extension along the midline of the robotic body and provides real-time proprioceptive feedback. The sensor is mounted such that its two ends rest above the front and tail cuffs, so the signal integrates the bending deformation over the full active body length. The sensor is narrower and softer than the silicone actuators and the backbone foil, so its contribution to the overall structural stiffness is negligible and is not modelled in the controller. In contrast to resistive eGaIn sensors used in earlier closed-loop prototypes on this platform family~\cite{Jusufi2017,Lin2021}, the capacitive element measures signed bending angles directly from a single midline-mounted unit, simplifying calibration and removing the need to pair sensors on opposite sides of the body. This direct, on-body measurement of the actual deformation is what enables more precise motion control under varying hydrodynamic loading: by exposing the body's true bending response — including the periodic component induced by the flow — at the $1~\mathrm{kHz}$ control-loop rate, the proprioceptive signal makes the flow-induced bias observable in the tracking error $\tilde q(t)$ and therefore correctable by both the stabilising feedback in equation~\eqref{eq:F} and the cycle-to-cycle update of the periodic reference input estimate $\hat{\xi}_{*}(t)$, neither of which an open-loop or purely model-based command can exploit.

\subsection{Real-time control hardware}
\label{sec:control_hw}

The control loop runs on a real-time microcontroller (myRIO-1900, National Instruments) that samples the bending sensor over I$^2$C at $1~\mathrm{kHz}$ and executes the PID-LRLES control law at the same rate. The microcontroller computes the bending-envelope estimate $q(t)$ from the raw sensor signal, evaluates the feedback signal $\mathcal{F}(t)$ in equation~\eqref{eq:F}, and outputs the analogue pressure command to the regulator together with the digital switching pattern for the solenoid valves. The closed-loop architecture realised on this hardware is summarised in Figure~\ref{fig:setup}(c).

\section{Learning Control Architecture}

The control problem is formulated on the bending-angle envelope $q(t)$ of the body--caudal undulation, which is reconstructed in real time from the embedded soft bending sensor. The controlled quantity is therefore not the instantaneous bending angle but its slowly varying amplitude envelope, while the underlying tail oscillation is sustained at a separately commanded flapping frequency. This formulation follows our previous static-water study~\cite{Schwab2024} and is retained here to enable a direct comparison under dynamic flow.

\subsection{Reference trajectory and feedback signal}
\label{sec:reference}

The desired bending-envelope trajectory is a periodic signal $q_{*}(t)$ of known period $T$, defined by an amplitude $A$, offset $q_0$ and phase $\phi_0$:
\begin{equation}
    q_{*}(t) \;=\; q_0 + A\sin\!\left(\tfrac{2\pi}{T}\,t + \phi_0\right).
    \label{eq:reference}
\end{equation}
Let $\tilde q(t) = q(t) - q_{*}(t)$ denote the tracking error. Following~\cite{tomei2015linear,verrelli2022pi,Schwab2024}, the controller is driven by the filtered feedback signal
\begin{equation}
    \mathcal{F}(t) \;=\; \dot{\tilde q}(t) + \gamma\,\tilde q(t),
    \label{eq:F}
\end{equation}
where $\gamma > 0$ weights the proportional component relative to the derivative one. The control objective is to drive $\mathcal{F}(t) \to 0$, which implies asymptotic tracking of the periodic reference~\eqref{eq:reference} over repeated undulation cycles.

\subsection{PID control baseline}
\label{sec:pid}

The baseline controller is the $\mathcal{F}$-PI law
\begin{equation}
    U_{\mathrm{PID}}(t) \;=\; -k_p\,\mathcal{F}(t) - k_i\int_{0}^{t}\mathcal{F}(\tau)\,\mathrm{d}\tau,
    \label{eq:pid_F}
\end{equation}
which, on substituting~\eqref{eq:F}, is equivalent to the conventional $\tilde q$-PID law
\begin{equation}
    U_{\mathrm{PID}}(t) \;=\; -K_p\,\tilde q(t) - K_i\int_{0}^{t}\tilde q(\tau)\,\mathrm{d}\tau - K_d\,\dot{\tilde q}(t),
    \label{eq:pid_q}
\end{equation}
with the gain mapping
\begin{equation}
    K_p = k_p\gamma + k_i,\quad K_i = k_i\gamma,\quad K_d = k_p.
    \label{eq:gain_map}
\end{equation}
The PID controller stabilises the error system and exactly regulates the output for constant references. For the periodic reference~\eqref{eq:reference}, however, the integral action cannot reconstruct a non-constant disturbance signal, so a residual tracking error persists across cycles. This residual is what the learning scheme below is designed to remove, acting on the estimate of the periodic reference input.

\subsection{Linear Repetitive Learning Estimation Scheme}
\label{sec:lrles}

Repetitive learning control exploits the periodicity of $q_{*}(t)$ by reinforcing the integral action of the PID law with cycle-to-cycle memory of the tracking error~\cite{xu2006repetitive,verrelli2022pi}. A finite-dimensional implementation suitable for real-time embedded control is obtained by approximating the pure delays $\mathrm{e}^{-isT}$, $i = 1, \ldots, p$, by $[m,m]$-Pad\'e expansions~\cite{verrelli2015linear,tomei2015linear}:
\begin{equation}
    \mathcal{P}_{[m,m]}(sT) \;=\; \frac{P_m(-sT)}{P_m(sT)},\quad P_m(sT) = \sum_{k=0}^{m}\binom{m}{k}\frac{(2m-k)!}{(2m)!}(sT)^{k}.
    \label{eq:pade}
\end{equation}
A weighted sum over the previous $p$ cycles, with positive weights $\alpha_{i}$ satisfying $\sum_{i=1}^{p}\alpha_{i} = 1$, defines the proper rational transfer function
\begin{equation}
    P(s) \;=\; \sum_{i=1}^{p}\alpha_{i}\,\mathcal{P}_{[m,m]}(isT) \;\doteq\; \frac{n_p(s)}{d_p(s)}.
    \label{eq:Ps}
\end{equation}
With $b = \sum_{i=1}^{p} i\,\alpha_{i}$, a learning gain $\mu > 0$ and a stabilising first-order filter $\beta/(\alpha s + 1)$, $\alpha \in [0,1)$, $\beta \in (0,1]$, the closed-form learning transfer function reads
\begin{equation}
    \mathcal{LC}(s) \;=\; \frac{\mathcal{L}[\hat{\xi}_{*}(t)](s)}{\mathcal{L}[\mathcal{F}(t)](s)} \;=\; \frac{-\mu\,b\,d_p(s)}{q_{\pi}(s)},\quad q_{\pi}(s) = (\alpha s + 1)d_p(s) - \beta\,n_p(s).
    \label{eq:LC}
\end{equation}
The filter parameters $\alpha$ and $\beta$ are chosen so that all roots of $q_{\pi}(s)$ lie in the open left half-plane, which guarantees that $\mathcal{LC}(s)$ is stable. The resulting $(1 + p\cdot m)$-dimensional state-space realisation generalises the classical integral action of the PID law from constant to periodic references while avoiding the long-term high-frequency-noise instability of na\"ive delay-based repetitive control schemes. With the values $p = 3$ and $m = 7$ used in this study (Table~\ref{tab:gains}), the learning filter has only $1 + p\cdot m = 22$ states and executes within a single $1~\mathrm{ms}$ control step on the myRIO-1900 target, so the architecture remains amenable to embedded deployment on low-power real-time hardware without GPU support.

\subsection{Closed-loop PID-LRLES integration}
\label{sec:pid_lrles}

The PID-LRLES control law adds the learning term in parallel with the PID block:
\begin{equation}
    \mathcal{L}[U_{\mathrm{PID\text{-}LRLES}}(t)](s) \;=\; -\left(k_p + \frac{k_i}{s} + \mathcal{LC}(s)\right)\mathcal{L}[\mathcal{F}(t)](s).
    \label{eq:pid_lrles}
\end{equation}
At each control step, $\mathcal{F}(t)$ is computed from the embedded soft bending-sensor signal through~\eqref{eq:F}, fed simultaneously into the stabilising PID block~\eqref{eq:pid_F} and into the finite-dimensional state-space realisation of $\mathcal{LC}(s)$, and the sum drives the pneumatic actuation pressure command, as summarised in the closed-loop block diagram of Figure~\ref{fig:setup}(c). Embedded proprioceptive feedback therefore enters both the stabilising and the learning paths, so that cycle-to-cycle updates of the periodic reference input estimate $\hat{\xi}_{*}(t)$ reflect the actual body deformation under the prevailing hydrodynamic loading rather than the open-loop actuation command. In static water, this architecture was shown to reduce envelope tracking error relative to a separately tuned PID baseline~\cite{Schwab2024}. The central question of the present work is whether the same advantage carries over when the hydrodynamic disturbance is no longer quasi-static, which we evaluate experimentally in Section~\ref{sec:results}.

\section{Experimental Setup}
\label{sec:setup}

\subsection{Recirculating flow tank}
\label{sec:flowtank}

Experiments were conducted in the recirculating water channel at Empa, with a fully transparent test section of cross-section $0.6 \times 1~\mathrm{m^2}$ extending up to $6~\mathrm{m}$ in length~\cite{Schwab2022}. A variable-speed pump regulates the bulk flow velocity over the range $0.02$--$1.5~\mathrm{m\,s^{-1}}$ and maintains stable, near-laminar flow in the test section. The robotic fish was suspended in the centre of the test section through the rigid mast described in Section~\ref{sec:body}, and stabilised laterally by two auxiliary aluminium beams attached to the mast. A high-speed camera placed below the test section recorded the midline kinematics at $400~\mathrm{frames\,s^{-1}}$, providing an optical reference for the soft-sensor-based bending-angle estimate. A six-axis force/torque sensor (ATI Nano17) mounted at the attachment point recorded the hydrodynamic loads transmitted from the fish to the mast.

\subsection{Reference trajectory and flow conditions}
\label{sec:trajectory}

The desired bending-envelope trajectory was a single sinusoid (equation~\eqref{eq:reference}) with period $T = 30~\mathrm{s}$, amplitude $A = 0.15~\mathrm{rad}$ and offset $q_0 = 0.6~\mathrm{rad}$, matching the AMAM conference protocol of~\cite{SchwabAMAM2025}. Underneath this envelope, the body was driven at a fixed flapping frequency $f_{\mathrm{flap}} = 1.2~\mathrm{Hz}$ with a duty cycle of $0.55$ and $5\%$ co-contraction between the antagonistic actuators, following~\cite{Schwab2024}.

The platform was evaluated at five bulk flow velocities, $U \in \{0,\; 5.10,\; 13.6,\; 22.7,\; 32.6\}~\mathrm{cm\,s^{-1}}$, spanning from quiescent water up to the upper speed at which body--caudal undulation can still be maintained against the incoming flow on this platform. The corresponding Reynolds numbers $\mathrm{Re} = U L_b / \nu$, with the body length $L_b \approx 0.23~\mathrm{m}$ from Section~\ref{sec:body} and water kinematic viscosity $\nu \approx 1.0 \times 10^{-6}~\mathrm{m^2\,s^{-1}}$, range from $\mathrm{Re} = 0$ at quiescent water up to $\mathrm{Re} \approx 7.5 \times 10^{4}$ at the highest speed.

\subsection{Controller tuning}
\label{sec:tuning}

The PID baseline and the PID-LRLES were each tuned separately in static water following the manual trial-and-error procedure of~\cite{Schwab2024}. The resulting gains, summarised in Table~\ref{tab:gains}, were kept fixed across all flow conditions reported in Section~\ref{sec:results}. Holding the gains fixed isolates the contribution of the cycle-to-cycle learning term $\mathcal{LC}(s)$ from any adaptation in the linear feedback path.

\begin{table}[htbp]
    \centering
    \caption{Controller gains used in the dynamic-flow experiments, carried over without modification from the static-water companion study~\cite{Schwab2024}. Both controllers were tuned in static water at $f_{\mathrm{flap}} = 1.2~\mathrm{Hz}$. PID gains are reported in the $\tilde q$-PID form (equation~\eqref{eq:pid_q}); PID-LRLES gains follow the notation of Section~\ref{sec:lrles}.}
    \label{tab:gains}
    \begin{tabular}{lll}
        \toprule
        Controller & Parameter & Value \\
        \midrule
        \multirow{3}{*}{PID} & $K_p$ & $3$ \\
                             & $K_i$ & $0.28$ \\
                             & $K_d$ & $0.15$ \\
        \midrule
        \multirow{9}{*}{PID-LRLES} & $k_p$ & $0.01$ \\
                                   & $k_i$ & $0.3$ \\
                                   & $\gamma$ & $4$ \\
                                   & $\mu$ & $0.4950$ \\
                                   & $\alpha$ & $0$ \\
                                   & $\beta$ & $0.99$ \\
                                   & $p$ (cycles in memory) & $3$ \\
                                   & $m$ (Pad\'e order) & $7$ \\
                                   & $(\alpha_1, \alpha_2, \alpha_3)$ & $(0.3,\, 0.2,\, 0.5)$ \\
        \bottomrule
    \end{tabular}
\end{table}

\subsection{Experimental protocol and data analysis}
\label{sec:protocol}

Every trial followed the same two-phase protocol. (i) An \emph{open-loop warm-up} stage in which the antagonistic actuators were inflated alternately at fixed timing until the bending-angle envelope reached steady state. (ii) A \emph{closed-loop} stage of $30~\mathrm{s}$, corresponding to one full period of the reference envelope, during which the selected controller (PID or PID-LRLES) was activated and commanded to track the reference of Section~\ref{sec:trajectory}. For each flow condition, $n = 5$ independent repetitions of the closed-loop stage were recorded for each controller, with the order of PID and PID-LRLES trials interleaved to avoid systematic drift.

The bending-angle envelope $q(t)$ was extracted from the soft sensor signal by taking the cycle-wise maximum of the absolute bending angle, providing a positive, slowly varying signal suitable for envelope-tracking analysis. Tracking accuracy was quantified by the root-mean-square error
\begin{equation}
    \mathrm{RMSE} \;=\; \sqrt{\frac{1}{t_2 - t_1}\int_{t_1}^{t_2} \tilde q^2(t)\, \mathrm{d}t},
    \label{eq:rmse}
\end{equation}
evaluated over the steady-state window $[t_1, t_2] = [10, 30]~\mathrm{s}$, which excludes the initial transient and covers two thirds of the reference period. Across-trial variability was characterised by the standard deviation and interquartile range of the per-trial RMSE within each flow condition. The PID and PID-LRLES distributions of per-trial RMSE were compared with a paired Wilcoxon signed-rank test at significance level $0.05$, matching the analysis previously reported in~\cite{SchwabAMAM2025}.

\section{Results}
\label{sec:results}

\subsection{Trajectory tracking under dynamic flow}
\label{sec:tracking}

\begin{figure}[!ht]
    \centering
    \includegraphics[width=\linewidth]{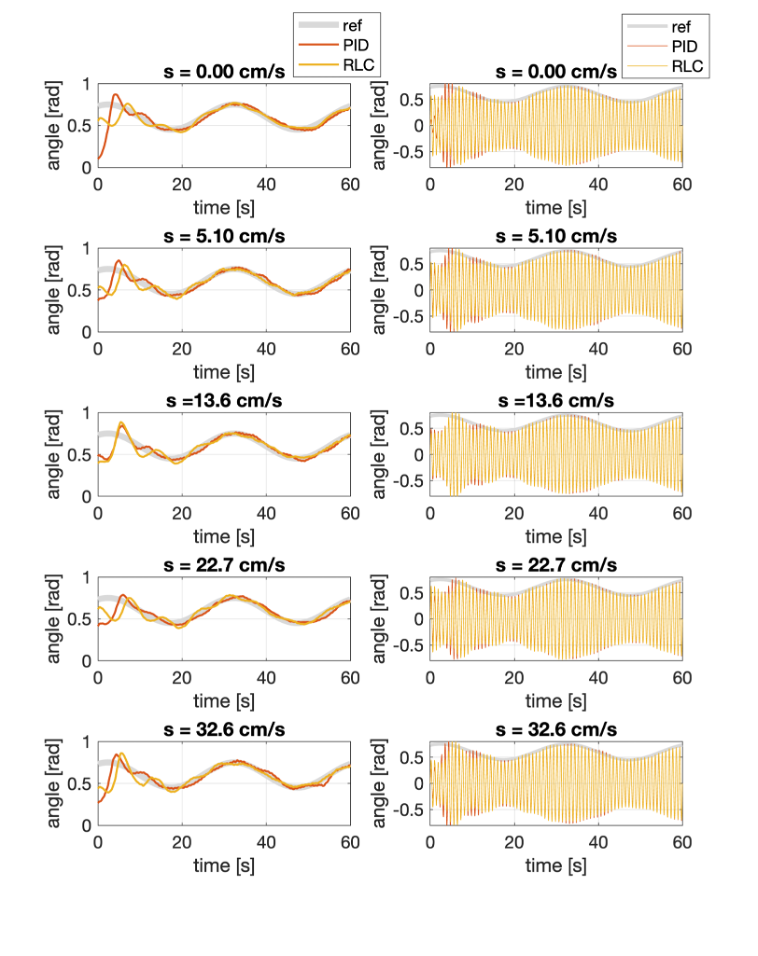}
    \caption{
    Dynamic-flow tracking performance of the soft robotic fish under conventional PID control and PID-LRLES control at five bulk flow speeds $U \in \{0, 5.10, 13.6, 22.7, 32.6\}~\mathrm{cm\,s^{-1}}$. Each panel shows the measured bending angle and reconstructed envelope against the desired reference $q_{*}(t) = 0.6 + 0.15\sin(2\pi t / 30)~\mathrm{rad}$. PID-LRLES maintains closer alignment with the reference trajectory and exhibits reduced tracking delay as flow speed increases. The $U = 0~\mathrm{cm\,s^{-1}}$ traces are reproduced from~\cite{Schwab2024} (Figures~7 and~8) under the CC~BY~4.0 licence; the remaining four flow speeds are new data presented here for the first time.
    }
    \label{fig:dynamic_flow_tracking}
\end{figure}

Figure~\ref{fig:dynamic_flow_tracking} compares the measured bending angle and the reconstructed envelope $q(t)$ against the reference $q_{*}(t)$ for both controllers, at each of the five flow speeds described in Section~\ref{sec:trajectory}. Conventional PID feedback maintains stable periodic oscillation across the full range of flow conditions, but its envelope tracks the reference with a visibly growing delay and amplitude deficit as flow speed increases: at $U = 0~\mathrm{cm\,s^{-1}}$ the peak envelope deviation from the reference is approximately $0.02~\mathrm{rad}$, while at $U = 32.6~\mathrm{cm\,s^{-1}}$ the deviation reaches approximately $0.04~\mathrm{rad}$.\footnote{The peak-deviation, convergence-time and residual-band values quoted in Section~\ref{sec:tracking} were read from the time-domain envelope panels of Figure~\ref{fig:dynamic_flow_tracking} and represent indicative magnitudes rather than independent measurements; the corresponding per-trial RMSE values reported in Table~\ref{tab:rmse} provide the quantitative comparison.} This degradation reflects the inability of fixed-gain feedback to anticipate the periodic component of the hydrodynamic load on the compliant body.

PID-LRLES tracks the same reference more closely under all five flow conditions. The learning term $\mathcal{LC}(s)$ progressively refines the actuation command from cycle to cycle within each $30~\mathrm{s}$ trial, so that after approximately $10~\mathrm{s}$ of closed-loop operation the envelope lies within $\pm 0.015~\mathrm{rad}$ of the reference even at the highest flow speed. The qualitative behaviour generalises the static-water observation of~\cite{Schwab2024} to dynamic flow and confirms the trend reported in our preliminary conference study~\cite{SchwabAMAM2025}.

Closer inspection of the envelope panels in Figure~\ref{fig:dynamic_flow_tracking} reveals low-amplitude bumps superposed on the slowly varying envelope under both controllers, whose magnitude grows with flow speed. We do not interpret these as control artefacts. Equivalent local modes were already visible in the midline-kinematics analysis of the static-water companion study (Figure~5 of~\cite{Schwab2024}) and are most plausibly read as signatures of the nonlinear coupling between the compliant body and the surrounding fluid: bifurcation-like local modes that arise from body--fluid interaction terms outside the periodic-reference structure that the learning estimator is designed to absorb. Their persistence across both controllers, and their amplitude scaling with $U$, are consistent with this interpretation rather than with residual learning transients of the PID-LRLES filter.

\subsection{Tracking-error variability}
\label{sec:variability}

To quantify control repeatability, the per-trial RMSE defined in equation~\eqref{eq:rmse} was computed over the steady-state window $[10, 30]~\mathrm{s}$ for each of the $n = 5$ repetitions at each flow speed. Per-condition mean values are summarised in Table~\ref{tab:rmse}, with the lower-error controller at each flow speed highlighted in bold. Across all five flow conditions PID-LRLES produces tracking errors that are more tightly concentrated around zero than those of the PID baseline, and the per-condition mean RMSE is smaller for PID-LRLES at every flow speed.

\begin{table}[!ht]
    \centering
    \caption{Per-condition mean RMSE of the bending-envelope tracking error, evaluated over $[10, 30]~\mathrm{s}$ across $n = 5$ trials per condition; per-trial RMSE variability is shown in Figure~\ref{fig:error_statistics}(b). Bold entries indicate the lower-error controller at each flow speed. Mean values are estimated from the per-sample error distributions reported in~\cite{SchwabAMAM2025}, Figure~1(a).}
    \label{tab:rmse}
    \begin{tabular}{lcc}
        \toprule
        $U$ [cm\,s$^{-1}$] & RMSE$_{\mathrm{PID}}$ [rad] & RMSE$_{\mathrm{PID\text{-}LRLES}}$ [rad] \\
        \midrule
        $0$    & $0.025$ & $\mathbf{0.015}$ \\
        $5.10$ & $0.022$ & $\mathbf{0.012}$ \\
        $13.6$ & $0.025$ & $\mathbf{0.013}$ \\
        $22.7$ & $0.029$ & $\mathbf{0.014}$ \\
        $32.6$ & $0.034$ & $\mathbf{0.018}$ \\
        \bottomrule
    \end{tabular}
\end{table}

Two further patterns are visible in the per-condition means of Table~\ref{tab:rmse}. The PID baseline shows a monotonic increase in mean RMSE with flow speed (from $0.025~\mathrm{rad}$ at $U = 0$ to $0.034~\mathrm{rad}$ at $U = 32.6~\mathrm{cm\,s^{-1}}$), consistent with the growing tracking delay observed in Figure~\ref{fig:dynamic_flow_tracking}: as the periodic hydrodynamic load on the compliant body grows with flow speed, the fixed-gain PID controller has no mechanism to anticipate it and the residual error grows accordingly. PID-LRLES exhibits a much flatter dependence on flow speed (mean RMSE rises only from $0.015$ to $0.018~\mathrm{rad}$ over the same range), indicating that the cycle-to-cycle memory-based estimate $\hat{\xi}_{*}(t)$ absorbs the bulk of the periodic disturbance regardless of its amplitude. The reduction in inter-trial spread is the key behaviour-level signature of the learning controller: by estimating the recurring component of the hydrodynamic disturbance through $\hat{\xi}_{*}(t)$, the controller removes the trial-to-trial drift that fixed-gain PID cannot compensate.

\begin{figure}[!ht]
    \centering
    \includegraphics[width=\linewidth]{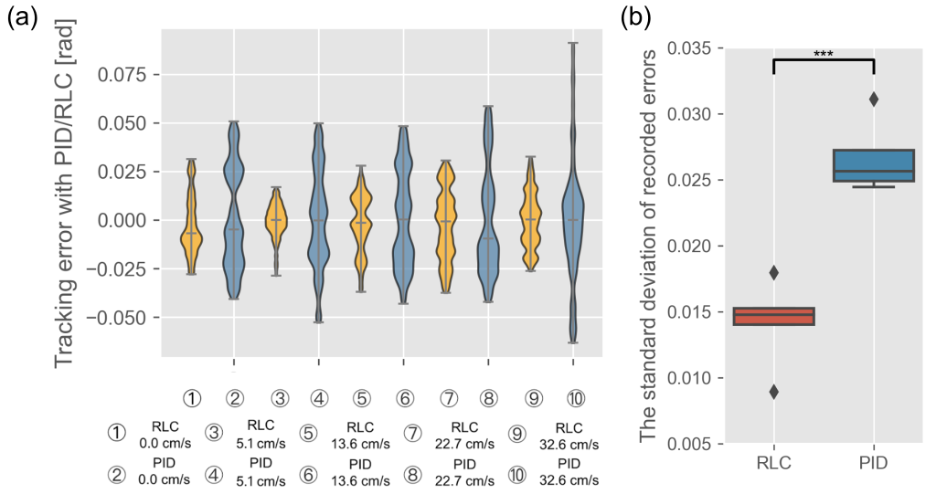}
    \caption{
    Tracking-error statistics under varying flow conditions. (a) Distributions of the per-sample tracking error $\tilde q(t)$ over the steady-state window for each controller at each flow speed; PID-LRLES errors are more tightly concentrated around zero. (b) Per-trial RMSE pooled across flow speeds; PID-LRLES significantly reduces the spread of the per-trial error ($n = 5$ per condition, paired Wilcoxon, $p = 1.8 \times 10^{-4}$). Both panels are reproduced from our preliminary AMAM~2025 conference contribution~\cite{SchwabAMAM2025} with permission of the authors.
    }
    \label{fig:error_statistics}
\end{figure}

To test whether this advantage is statistically supported, we applied a paired Wilcoxon signed-rank test on the per-trial RMSE distributions, pooled across the five flow conditions ($n_{\mathrm{pairs}} = 25$). PID-LRLES significantly reduces tracking-error variability compared with the PID baseline ($p = 1.8 \times 10^{-4}$), reproducing the effect size previously reported in~\cite{SchwabAMAM2025}. The corresponding error distributions are shown in Figure~\ref{fig:error_statistics}: panel~(a) compares the per-sample tracking-error distributions condition by condition, and panel~(b) pools the per-trial RMSE across all five flow speeds for the two controllers, providing a direct visualisation of the inter-trial variability that the Wilcoxon test summarises.

The PID-LRLES distributions in Figure~\ref{fig:error_statistics}(a) are not strictly unimodal: each violin shows low-amplitude side lobes whose prominence grows with flow speed. These bumps are the per-sample manifestation of the same nonlinear body--fluid coupling signature discussed for the time-domain envelopes in Section~\ref{sec:tracking}. Because the learning estimator has absorbed the dominant periodic disturbance, what remains visible in the residual error distribution is the bifurcation-like local modes of the compliant-body-fluid interaction, which are outside the periodic-reference structure that $\mathcal{LC}(s)$ is designed to track. The equivalent multi-lobe signature is documented in the midline-kinematics analysis of the static-water companion study (Figure~5 of~\cite{Schwab2024}), where the body adopts characteristic asymmetric shapes at intermediate envelope amplitudes; the bumps in Figure~\ref{fig:error_statistics}(a) are the per-sample error-distribution image of the same body--fluid bifurcation, observed here under flow.

\subsection{Robustness under hydrodynamic disturbances}
\label{sec:robustness}

The flow-speed dependence summarised in Section~\ref{sec:variability} has an interpretation that goes beyond per-condition accuracy. A controller that performs slightly better on average but whose performance degrades sharply as flow speed grows is of limited use for comparative robophysical experiments, in which the same platform must be operated reproducibly across a range of hydrodynamic conditions. The relevant operational measure of robustness is therefore not the mean RMSE at any single flow speed but the sensitivity of the controller's performance envelope to the prevailing flow condition. Under this measure, PID-LRLES is markedly more robust than the PID baseline: as evident in Figure~\ref{fig:error_statistics}(a), the per-sample tracking-error distribution widens substantially with flow speed under PID, whereas under PID-LRLES the distribution remains tightly concentrated around zero at every flow condition. Learning control therefore not only improves tracking accuracy at any individual flow speed but also flattens the controller's sensitivity to the prevailing hydrodynamic condition --- the operational definition of robustness relevant to robophysical swimming experiments.

\section{Discussion}

\subsection{Trajectory tracking and variability under dynamic flow}
\label{sec:disc_tracking}

The dynamic-flow experiments reproduce, in a hydrodynamically loaded setting, the qualitative advantage of PID-LRLES over conventional PID that we previously reported in static water~\cite{Schwab2024}, and they extend the preliminary observation of~\cite{SchwabAMAM2025} to a statistically supported result. Across the five tested flow speeds, the PID baseline maintains stable periodic oscillation but accumulates a growing tracking delay and amplitude deficit as flow speed increases (Figure~\ref{fig:dynamic_flow_tracking}), while PID-LRLES tracks the same reference more closely after a transient of approximately $12$ flapping cycles, consistent with the $\sim 10~\mathrm{s}$ convergence time at $f_{\mathrm{flap}} = 1.2~\mathrm{Hz}$. The reduction in per-trial RMSE variability is significant in the pooled comparison ($p = 1.8 \times 10^{-4}$, $n_{\mathrm{pairs}} = 25$) and Table~\ref{tab:rmse} shows that the gain is largest at the higher flow speeds, consistent with the interpretation that the learning term $\mathcal{LC}(s)$ contributes most when the periodic component of the hydrodynamic disturbance is largest.

It is worth noting that the controller gains were tuned once, in static water, and held fixed across the full range of flow conditions (Table~\ref{tab:gains}). The PID baseline therefore cannot adapt to changes in the apparent fluid stiffness of the compliant body--fluid system, whereas PID-LRLES recovers an equivalent integral action for the periodic disturbance through the Pad\'e-approximated memory of the previous $p = 3$ cycles. The advantage demonstrated here is therefore not the result of separate tuning at each flow speed but of the structural property that the learning term reconstructs the periodic component of the disturbance from past errors.

\subsection{Role of embedded soft proprioceptive feedback}
\label{sec:disc_sensor}

The closed-loop architecture is meaningful only insofar as the controller can observe the actual deformation of the compliant body. The capacitive soft bending sensor (Section~\ref{sec:sensor}) supplies this observation at the $1~\mathrm{kHz}$ control-loop rate, so that both the stabilising feedback signal $\mathcal{F}(t)$ in equation~\eqref{eq:F} and the cycle-to-cycle update of the estimate $\hat{\xi}_{*}(t)$ in $\mathcal{LC}(s)$ are driven by the true body response under the prevailing hydrodynamic load rather than by the open-loop actuation command. In an open-loop or model-based architecture, the periodic component of the hydrodynamic disturbance would manifest as an unmodelled bias in the bending envelope that the controller has no mechanism to remove; with embedded proprioception in the loop, the same bias appears in $\tilde q(t)$, is absorbed into the memory term over a few cycles, and is then cancelled in the actuation command. Soft sensing and repetitive learning are therefore complementary rather than independent contributions to robustness.

\subsection{Implications for robophysical swimming studies}
\label{sec:disc_robophysics}

The broader value of improved control repeatability lies in robophysical inference. Recent comparative studies on this family of soft robophysical fish platforms have used them as robophysical models to test hypotheses about morphology-dependent swimming dynamics: differences in pitch torque and thrust across extinct ichthyosauriform caudal-fin shapes~\cite{Sprumont2024} and the hydrodynamic signature of body--caudal undulation under varying flow~\cite{Schwab2022} are both interpretations that rely on the assumption that observed differences arise from morphology rather than from variability in the actuation envelope. Figures~\ref{fig:error_statistics}(a) and~\ref{fig:error_statistics}(b) make explicit what was previously implicit: under fixed-gain PID the tracking-error distribution broadens markedly as flow speed increases and the per-trial RMSE remains significantly more dispersed than under PID-LRLES, whose error envelope is comparatively flat across flow conditions. Reducing this flow-dependent control-induced variability strengthens the case that morphology-dependent differences in thrust or pitch torque measured on the platform reflect the morphology of the appended fin rather than residual variability in the body--caudal kinematics. In this sense, learning-enhanced closed-loop control acts as an enabling layer for future morphology-dependent robophysical experiments, including extensions of~\cite{Sprumont2024} to a wider range of flow speeds and to fin shapes whose performance differences may be subtler than the controller's residual envelope error.

A second confound has to be acknowledged when the appended fin is asymmetric about the body midline, as in the heterocercal ichthyosauriform morphologies of Figure~\ref{fig:fins_torque} and~\cite{Sprumont2024}. Such fins generate vertical and rotational loads that are not symmetric about the actuation plane, and these loads can induce torsional deformation of the compliant body in addition to the desired pitch torque at the mast attachment. The pitch-torque signal measured at the mast therefore reflects both the intended fin-shape contribution and a passive body-torsion response, and disentangling the two is a separate question that the present envelope-tracking framework does not address. A useful complement to the envelope-tracking accuracy reported here would therefore be a kinematic measurement of body torsion under each fin shape, which would let comparative pitch-torque differences be attributed to fin morphology rather than to fin-induced body deformation.

\subsection{Towards motor-command tuning of body mechanics}
\label{sec:disc_mct}

A further direction opens up when the present results are read against the embodied-intelligence framework of~\cite{Nanayakkara2024}, which separates biological motor commands into two complementary classes: motor commands that directly drive movement and interaction forces (MCM), and motor commands that tune body mechanics --- limb stiffness, joint impedance, body geometry --- to suit the task (MCT). The efficiency of the former is determined by how well the latter is regulated, and soft-bodied robots provide a particularly clean setting in which to dissect this coupling. The PID-LRLES controller evaluated here is squarely an MCM-class controller: it modulates the actuation pressure command in closed loop while the mechanical impedance of the body --- backbone stiffness, fin geometry, baseline actuator pre-charge --- is fixed by hardware before each trial. The complementary MCT axis is, on the present platform, currently absent.

Two observations from our experiments suggest that adding an MCT degree of freedom is a natural next step. First, the residual flow-speed dependence of the PID-LRLES RMSE (Table~\ref{tab:rmse}, Figure~\ref{fig:error_statistics}) cannot in principle be removed by faster cycle-to-cycle learning alone if the underlying body--fluid dynamics themselves shift with flow speed; matching the body's apparent fluid stiffness to the prevailing hydrodynamic regime would attack the disturbance at its source rather than compensating for it after the fact. Second, recent demonstrations on related soft fish platforms have shown that real-time hydraulic stiffness modulation of the caudal fin can be used to improve thrust and propulsive efficiency~\cite{Obayashi2025}, providing a concrete experimental route to MCT on this class of robot.

Repetitive learning makes the composition of MCT and MCM particularly attractive on this platform. Holding the MCM-side controller (PID-LRLES) stable while an MCT-side command varies the apparent body stiffness during locomotion provides a clean way to isolate the morphology-by-flow term that is otherwise confounded by control-induced variability: any change in tracking residual, thrust or pitch torque can be attributed to the stiffness command rather than to drift in the underlying kinematic envelope. This sets up a natural follow-up programme on this and related platforms in which the same fin morphologies considered in~\cite{Sprumont2024} are tested across flow speeds under jointly active MCM and MCT control.

\subsection{Shape-driven Strouhal-number modulation as a near-term research programme}
\label{sec:disc_strouhal}

A direct consequence of fixing the flapping frequency in the present campaign is that the propulsive Strouhal number $\mathrm{St} = f_{\mathrm{flap}}\,A_{\mathrm{tail}}/U$, which biological steady swimmers concentrate in the efficient range $\mathrm{St} \in [0.2, 0.4]$~\cite{Taylor2003}, is not held at a target value but varies passively as $U$ is swept. We previously characterised the Strouhal-dependent thrust of an earlier version of this platform using 2D particle image velocimetry combined with dynamic mode decomposition (DMD) and showed that wake modes at integer multiples of $f_{\mathrm{flap}}$ carry the propulsive impulse~\cite{Schwab2022}. A natural follow-up programme is therefore to use the present learning controller as an actuation-rhythm regulator while the caudal-fin morphology --- not the flapping kinematics --- is varied to actively control St.

Two morphological levers are particularly natural to consider on this platform, both already represented in the asymmetric ichthyosauriform fin set of Figure~\ref{fig:fins_torque} and~\cite{Sprumont2024}. The first is the caudal-fin aspect ratio $\mathrm{AR} = s^2/A_{\mathrm{fin}}$ (span squared over planform area). At fixed flapping kinematics, varying AR rescales the effective peak-to-peak wake amplitude $A_{\mathrm{tail}}$ entering the Strouhal definition and therefore translates the propulsive operating point along the St axis at constant $f_{\mathrm{flap}}$ and $U$: high-AR lunate fins concentrate momentum into a narrow, high-efficiency wake, while low-AR paddle-like fins broaden it and shift the operating point upward in St. The second is the heterocercal asymmetry between the upper and lower lobes of the caudal fin. Breaking reflection symmetry reorganises the vortex shedding into an asymmetric reverse-Karman pattern in which the lobe contributions to the propulsive impulse no longer average symmetrically over a flapping cycle, redefining the dynamically relevant $A_{\mathrm{tail}}$ through a wake-weighted rather than geometric measure; this same asymmetry simultaneously produces the pitch-torque component reported in~\cite{Sprumont2024}, so that St modulation and morphology-dependent pitch torque are coupled through a single asymmetry parameter.

Taken together, AR change and heterocercal change open a two-parameter morphology surface $(\mathrm{AR},\,\varepsilon_{\mathrm{hetero}})$ on which the propulsive operating point $(\mathrm{St},\,\text{pitch component})$ of the platform can be steered without changing the actuation rhythm. The DMD-based wake-mode framework already established for the platform in~\cite{Schwab2022} provides the natural experimental protocol for measuring this modulation: at each $(\mathrm{AR},\,\varepsilon_{\mathrm{hetero}})$ configuration, the dominant DMD-mode amplitudes and frequencies should redistribute in characteristic ways, giving a directly observable signature of shape-driven St control. Combining such fin-shape modulation with the present learning controller --- which keeps the actuation envelope reproducible across flow conditions while the morphology is varied --- would promote the Strouhal number from a passive diagnostic to a behaviour-level KPI, and would let the morphology-dependent comparisons initiated in~\cite{Sprumont2024} be carried out under controlled, target-St operating conditions.

\subsection{Computational footprint and embedded deployment}
\label{sec:disc_compute}

A complementary advantage of the linear repetitive learning architecture used here is its modest computational footprint, which is in marked contrast to the prevailing trend in learning-based soft-robot control. The state-of-the-art in this area is dominated by data-driven neural-network policies whose training and on-line inference typically require GPU-class compute and are not straightforwardly embedded on the platform~\cite{Laschi2025}. PID-LRLES occupies the opposite end of this spectrum: the closed-form transfer function $\mathcal{LC}(s)$ in equation~\eqref{eq:LC} admits a $(1 + p\cdot m)$-dimensional state-space realisation that, with the values $p = 3$, $m = 7$ used here, contains only $22$ states and executes within the $1~\mathrm{ms}$ control step of the myRIO-1900 real-time microcontroller (Section~\ref{sec:control_hw}). Because the learning happens entirely on-board at the $1~\mathrm{kHz}$ control-loop rate without any off-line training, no dataset, and no GPU support, the architecture is a natural fit for the embedded, power-constrained operating regime in which untethered or autonomously swimming soft robots will ultimately be deployed.

\subsection{Limitations and future work}
\label{sec:disc_limitations}

Several limitations of the present study are worth stating explicitly. First, the platform was tested at a single flapping frequency ($f_{\mathrm{flap}} = 1.2~\mathrm{Hz}$); the dependence of the learning-controller advantage on the ratio between the reference period $T$ and the flapping period $1/f_{\mathrm{flap}}$ remains to be characterised, as is the propulsive Strouhal-number programme set out in Section~\ref{sec:disc_strouhal}. Second, the body is held stationary in the flow tank by a rigid mast and the present experiments therefore do not test free-swimming behaviour, transient hydrodynamic events such as gust responses, or three-dimensional manoeuvring. Third, the controller gains were tuned in static water and held fixed across flow conditions; whether gain-scheduling, on-line gain adaptation or extensions of the LRLES memory length $p$ would further reduce the residual flow dependence of the per-trial RMSE remains an open question. Finally, the present analysis is restricted to envelope tracking; thrust, pitch torque and energetic efficiency, all of which the existing platform instrumentation can measure (Section~\ref{sec:flowtank}), are natural next-step performance metrics for assessing whether the kinematic repeatability reported here translates into more reliable hydrodynamic comparisons across morphologies and flow conditions.

\section{Conclusions}

Learning control improves the robustness and repeatability of soft robotic body--caudal undulation under dynamic flow. With a single static-water gain set held fixed across five flow speeds from $0$ to $32.6~\mathrm{cm\,s^{-1}}$, PID-LRLES tracked the periodic bending-envelope reference more closely than a separately tuned PID baseline and significantly reduced the inter-trial spread of the per-trial RMSE ($p = 1.8 \times 10^{-4}$, paired Wilcoxon, $n_{\mathrm{pairs}} = 25$). The result extends the static-water validation of~\cite{Schwab2024} to a hydrodynamically loaded setting and confirms the trend reported in our preliminary conference study~\cite{SchwabAMAM2025}. By reducing flow-dependent control-induced variability while requiring only static-water tuning, learning-enhanced closed-loop control acts as an enabling layer for repeatable robophysical swimming experiments and for future morphology-dependent comparative studies of aquatic locomotion on this and similar soft platforms.

\ack{
The authors thank all collaborators and technical staff Terence Fontana of Engineering Science Department at Empa who supported the development of the robotic fish platform and experimental flow-tank infrastructure.
}

\funding{
This research was supported by two Swiss national Science Foundation Grants Projects 127024 and 182638 (to A.J.).
}

\data{
Supporting information available at \url{https://github.com/ardianet1/SoftSensoRepetLearn}.
}

\section*{Code availability}
The real-time control software (PID and PID-LRLES implementations on the myRIO-1900 target), the data analysis scripts used to compute the per-trial RMSE statistics and the source code for Figures~\ref{fig:dynamic_flow_tracking} and~\ref{fig:error_statistics} are available at \url{https://github.com/ardianet1/SoftSensoRepetLearn} alongside the supporting data referenced above.

\bibliographystyle{unsrt}
\bibliography{references}

\end{document}